\documentclass[10pt,twocolumn,letterpaper]{article}

\usepackage{wacv}              % To produce the CAMERA-READY version
\usepackage{graphicx}
\usepackage{amsmath}
\usepackage{amssymb}
\usepackage{booktabs}
\usepackage{enumitem}

\usepackage[pagebackref,breaklinks,colorlinks]{hyperref}

\usepackage[capitalize]{cleveref}
\crefname{section}{Sec.}{Secs.}
\Crefname{section}{Section}{Sections}
\Crefname{table}{Table}{Tables}
\crefname{table}{Tab.}{Tabs.}

\def\wacvPaperID{1450} % *** Enter the WACV Paper ID here
\def\confName{WACV}
\def\confYear{2025}
\def\etal{{et.~al}\xspace}

\def\roi{ROI\xspace}

\def\gbcnet{GBCNet\xspace}

\def\vita{ViT-Adapter\xspace}
\def\myarch{LQ-Adapter\xspace}

\newcommand{\myfirstpara}[1]{\par \noindent \textbf{#1.}}
\newcommand{\mypara}[1]{\vspace{0.1em} \myfirstpara{#1}}

\begin{document}

\title{LQ-Adapter: ViT-Adapter with Learnable Queries for Gallbladder Cancer Detection from Ultrasound Images}

% \author{Soumen Basu\textsuperscript{1}% 
% \thanks{~Soumen is currently affiliated to Samsung R\&D Institute Bangalore}~~\thanks{~Joint first authors}~, 
% Mayuna Gupta\textsuperscript{1~\textdagger}, Chetan Madan\textsuperscript{1}, Pankaj Gupta\textsuperscript{2}, Chetan Arora\textsuperscript{1} \\
% \textsuperscript{1} IIT Delhi \qquad 
% \textsuperscript{2} PGIMER, Chandigarh\\
% }

% \author{Mayuna \\
% \and

\author{Chetan Madan\textsuperscript{1}% 
\thanks{~Joint first authors}~, 
Mayuna Gupta\textsuperscript{1$\ast$}\thanks{~Currently affiliated to University of California San Diego}~, Soumen Basu\textsuperscript{1}\thanks{~Currently affiliated to Samsung R\&D Institute Bangalore}~, Pankaj Gupta\textsuperscript{2}, Chetan Arora\textsuperscript{1} \\
\textsuperscript{1} IIT Delhi \qquad 
\textsuperscript{2} PGIMER, Chandigarh\\
\url{https://github.com/ChetanMadan/LQ-Adapter}
}

\maketitle

%%%%%%%%% BODY TEXT
%%%%%%%%% ABSTRACT
\begin{abstract}
We focus on the problem of Gallbladder Cancer (GBC) detection from Ultrasound (US) images. The problem presents unique challenges to modern Deep Neural Network (DNN) techniques due to low image quality arising from noise, textures, and viewpoint variations. Tackling such challenges would necessitate precise localization performance by the DNN to identify the discerning features for the downstream malignancy prediction. While several techniques have been proposed in the recent years for the problem, all of these methods employ complex custom architectures. Inspired by the success of foundational models for natural image tasks, along with the use of adapters to fine-tune such models for the custom tasks, we investigate the merit of one such design, ViT-Adapter, for the GBC detection problem. We observe that ViT-Adapter relies predominantly on a primitive CNN-based spatial prior module to inject the localization information via cross-attention, which is inefficient for our problem due to the small pathology sizes, and variability in their appearances due to non-regular structure of the malignancy. In response, we propose, \myarch, a modified Adapter design for ViT, which improves localization information by leveraging learnable content queries over the basic spatial prior module. Our method surpasses existing approaches, enhancing the mean IoU (mIoU) scores by 5.4\%, 5.8\%, and 2.7\% over ViT-Adapters, DINO, and FocalNet-DINO, respectively on the US image-based GBC detection dataset, and establishing a new state-of-the-art (SOTA). Additionally, we validate the applicability and effectiveness of \myarch on the Kvasir-Seg dataset for polyp detection from colonoscopy images. Superior performance of our design on this problem as well showcases its capability to handle diverse medical imaging tasks across different datasets. Source code and trained models are publicly released.
\end{abstract}

\begin{figure}
    \centering
    \includegraphics[width=\linewidth]{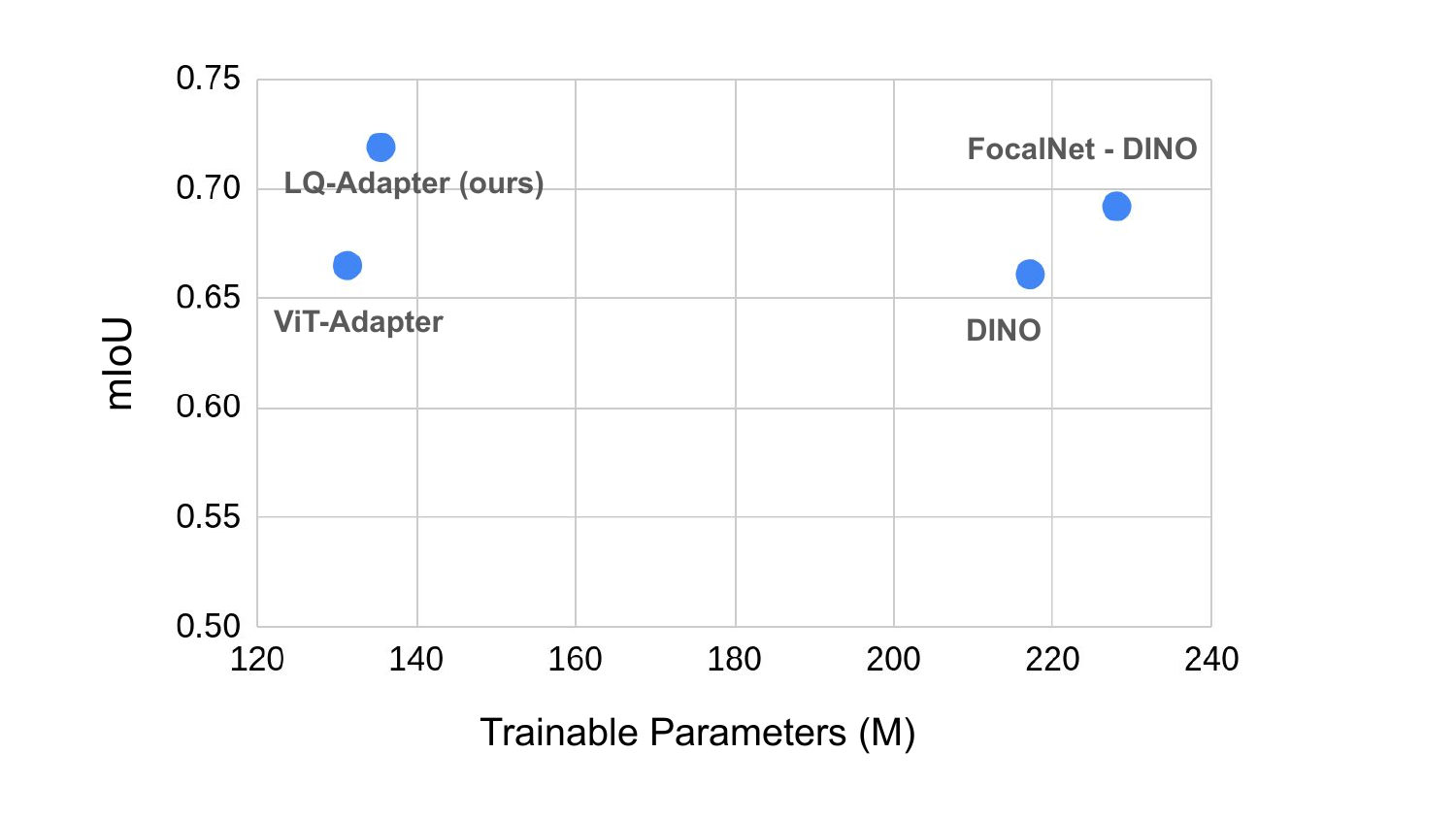}
    \caption{We compare model sizes and performance (mean intersection-over-union) of SOTA transformer-based object detection methods on the GBCU dataset. It highlights the superiority of \myarch by demonstrating its ability to achieve competitive performance while maintaining a more efficient parameter footprint than existing methods.}
    \label{fig:teaser}
\end{figure}
\section{Introduction}
\label{sec:intro}

% \subsection{Gallbladder cancer detection from ultrasound images}
%
Deep learning-based gallbladder cancer (GBC) detection from ultrasound (US) images has piqued researchers' interest in recent years. US image analysis poses several challenges, such as low image quality due to noise, artifacts like shadows or echogenic textures, and viewpoint variation due to handheld sensors. US images of the gallbladder are challenging to use in deep neural networks (DNNs) due to high intra-class (variability in view due to the 2D slicing of a 3D organ) and low inter-class variability (GBC typically occupies a very small portion of the image) \cite{basu2023gall}. 

Several efforts have been made in the literature to address the GBC detection from US images \cite{basu2022surpassing,basu2023radformer,basu2022unsupervised,gbc-lancet,gbc-xgc}. Basu \etal proposed \gbcnet \cite{basu2022surpassing}, a two-stage design that initially generates the regions-of-interest via a FasterRCNN-based detector, and then uses a specialized classifier called MS-SoP to classify these regions. Using such focused regions help to retain the crucial malignant features and mitigate the effect of noise and artifacts in the US images. RadFormer~\cite{basu2023radformer} uses a global-local attention and bag-of-words style feature embedding on locally focused regions to achieve SOTA GBC detection performance. However, these methods are primarily custom architectures and are not easily applicable to other datasets or tasks. 

% \begin{figure*}[t]
%     \centering
%     \includegraphics[width=\textwidth]{figs/teaser.pdf}
%     \caption{We propose a Learnable Query-based Adapter (\myarch) for Gall Bladder Detection. With a minimalistic design \myarch can modify ViT\cite{vit} suitable for medical object detection, addressing various challenges like viewpoint variations, high intra-class variability and low inter-class variability}
%     \label{fig:teaser}
% \end{figure*}

 Transformer-based object detection has evolved through various approaches. Methods such as Detection Transformer (DETR) \cite{detr} and its variants like DINO\cite{dino} and Focal-DINO \cite{focalnet, focaldino} have been tailored for employing transformers towards detection tasks, using intricate design elements on top of the transformer backbones. Alternatively, architectures like \cite{vitdet, swin} modify transformers themselves, introducing scale or hierarchical elements to ameliorate their ability to link finer details with broader context.

 Lately, with the rising popularity of techniques like \cite{lora, adapters, adapterog} etc., fine-tuning of foundational models for fundamental vision tasks in different data scenarios has gained momentum. Along those lines works like \vita \cite{vita} have demonstrated great promise in leveraging frozen pre-trained backbones on relatively small datasets and achieving state-of-the-performance. Yet, in our analysis, we found \vita's primitive spatial prior module insufficient for capturing low-level details in medical image datasets especially in the context of GBC which manifests small sized object (pathology) with variable visual appearance. 
 %However, DETR variant models entail complex architectural interventions.  
% update with some rehash of related works 

\mypara{Contributions} 
The key contributions of this work are:
\begin{enumerate}[label=\textbf{(\arabic*)}]
\item We design a novel adapter -- \myarch to improve the primitive spatial prior module presented by ViT-Adapter and obtained 5.4\% improvement on the mean IoU score over \vita on GBC detection. Our proposed design also surpasses the DINO and FocalNet-DINO\cite{focaldino}, the current DETR-based SOTA by 5.8\%, 2.7\% respectively, in terms of mean IoU for GBC detection.
\item We also use the \roi generated by \myarch in the first stage of the \gbcnet architecture \cite{basu2022surpassing}, and obtain 93.4\% GBC classification accuracy, which outperforms RadFormer \cite{basu2023radformer}, the current SOTA, and the original \gbcnet with FasterRCNN-based \roi generation.   
\item \myarch is also the first attempt of using foundational models for GBC detection from US images. Instead of performing complex architectural interventions, a tunable lightweight adapter on top of a ViT based backbone is shown to be equally effective for GBC detection.
\item We also experimentally demonstrate the applicability of \myarch on DDSM dataset \cite{ddsm} for detecting breast lesions in mammography, which indicates the general applicability of \myarch to detect diverse types of cancers across different diagnostic imaging modalities.
\end{enumerate}

\section{Related works}
\subsection{Deep Learning for GB Abnormalities}
% Other 
While Deep Neural Networks (DNNs) have been explored for various gallbladder diseases, GBC detection using AI remains an active area of research \cite{gupta2024applications}. One initial line of works include Chang \etal \cite{chang2022ct} employing a UNet-based denoising technique to enhance the image quality of Low-Dose CT scans to characterize GBC. 
In the realm of ultrasound (US) imaging, researchers have employed a novel multi-scale second-order pooling (MS-SoP) CNN architecture with curriculum learning for efficient GBC detection  \cite{basu2022surpassing}. \cite{gbc-lancet} further studied the performance of MS-SoP in classifying different sub-types of GBC on a large prospective patient cohort. \cite{basu2022unsupervised} later utilized unsupervised contrastive learning to learn malignancy representations. 

On the other hand, \cite{basu2023radformer} exploits a transformer-based dual-branch architecture for accurate and explainable GBC detection. Transformer applications have extended beyond GB detection with studies investigating GBC differentiation from xanthogranulomatous cholecystitis \cite{xgc} and proposing weakly supervised detection methods using DETR \cite{basu2023gall}. Most recently, \cite{focusmae} was introduced, which uses spatial priors to mask selective regions in ultrasound videos, improving representation learning on GBC video datasets. Recognizing the challenges of limited data, researchers have developed calibration techniques for models trained on smaller datasets
\cite{gupta2023reliable}. This rich tapestry of research emphasizes the ongoing pursuit of robust AI methods for GBC detection. Our work builds upon this foundation by introducing a novel approach using learnable query-based adapters for GBC detection in US images. 

\subsection{Transformers for Object Detection}
DETR~\cite{detr} pioneered Transformer-based object detection, replacing hand-crafted components with an end-to-end approach. Despite its innovation, DETR faces challenges in convergence speed and small object detection. Deformable-DETR \cite{deformable}, DN-DETR \cite{dn}, DAB-DETR \cite{dab}, and DINO \cite{dino, dinodetr} address these issues through iterative refinement, dynamic anchor boxes, denoising, and mixed query selection. DAB-DETR introduced dynamic anchor boxes as content queries in DETR object queries. DINO enabled learnable content queries at decoder, enhancing spatial priors without encoder bias. Several newer object detection baseline  While DINO and FocalNet-DINO \cite{focalnet, focaldino,focalstabledino} achieves state-of-the-art performance on COCO, such DETR variants struggle with generalization to smaller medical datasets, and overfitting challenges.

\subsection{ViT Adapter}
Usually adapter belong to the parameter efficient fine tuning strategies. Low rank adaptation \cite{lora}, or weight decomposed low rank adapters \cite{dora} are some of the popular adapter techniques. ViT-Adapter \cite{vita} offers an alternative to DETR variants for localization tasks by utilizing a CNN-based spatial prior module to integrate localization information through cross-attention. It enables a pre-trained ViT to handle detection tasks with only task-specific tuning of the adapter component, avoiding the need for new architectures or re-training.  However, our analysis reveals that the spatial prior module in \vita is optimal only for natural images and dilutes local spatial information. Drawing inspiration from object detection models, we introduce a novel Adapter design, \myarch, which employs learnable content queries to enhance localization information within the Adapters, and improve the detection performance on medical imaging tasks such as GBC detection.

\begin{figure*}[t]
    \centering
    \includegraphics[width=\textwidth]{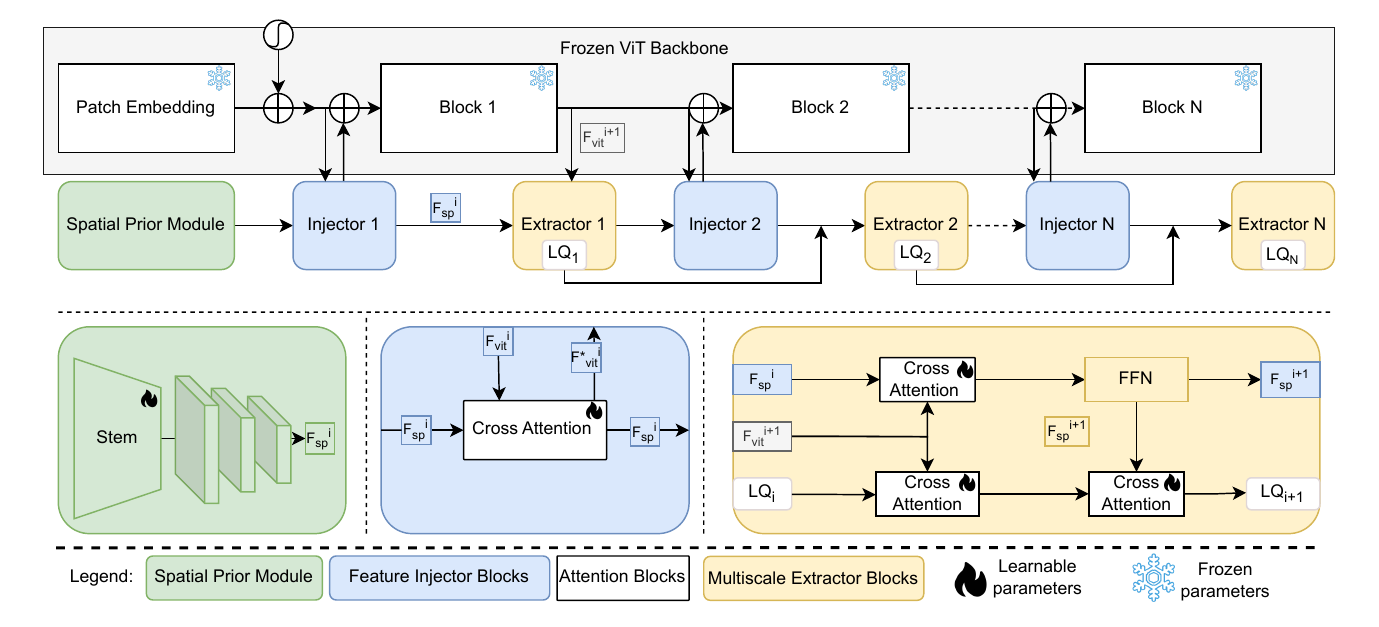}
    % update
    \caption{Schematic architecture diagram of the proposed \myarch. The learnable content queries are added to the extractor blocks of adapter modules for improved localization. FFN: Feed Forward Network, LQ: Learnable Queries, $F_{sp}$: Features from the Spatial Prior Module, $F_{vit}$: Features from the frozen ViT \cite{vit} backone}.
    \label{fig:arch}
\end{figure*}

\subsection{Learnable Query}

Learnable query-based refinement methods have demonstrated a significant advantage in enhancing model performance and adaptability. As evidenced by several recent studies \cite{bao2021beit,Shehzadi2024SparseSS, Xiong2024MultimodalLQ,dab, dino}, these techniques leverage the coupling of low-level features from the pre-trained backbones with high-level features derived from crude spatial features to generate enriched features. This integration facilitates more effective information extraction and modelling of complex relationships within data. Moreover, learnable queries enable neural networks to adjust their weights dynamically during refinement, optimizing query strategies based on learned patterns and contexts within the dataset. This approach could prove to be beneficial for In medical imaging tasks, enabling models to focus on nuanced low-level features critical for accurate diagnosis.

\subsection{Transfer Learning for Medical Image Analysis}

With the rising popularity of foundational models for natural images, several studies have been developed to adapt them for medical image analysis. Within vision, these can be classified into two types: Large Visual Models (LVMs), and Large Multi-modal models (LMMs). 

\mypara{Large Visual Models (LVMs)}
These models are pre-trained on massive datasets of natural images and then fine-tuned for specific datasets on imaging tasks like segmentation, classification, and detection. Notably, medical image segmentation has seen significant progress with LVMs, as evidenced by works like \cite{samadapter, medsam,huang2024segment, melo, xin2024parameter} etc. These approaches leverage transfer techniques like adapter modules, low-rank adaptation and prompt tuning\cite{melo, adapters, eft , lora, peftmia,univs} to adapt the pre-trained LVM to the specific characteristics of medical images, such as limited data and presence of noise or artifacts.

\mypara{Large Multi-modal Models (LMMs)}
This emerging class of foundational models goes beyond visual data. LMMsa are trained on a combination of modalities like images, text reports and electronic health records. This allows them to learn richer representations that incorporate visual information and relevant clinical context.\cite{clip,univs, pmlr,medclip} etc. LMMs hold great promise for tasks like diagnosis prediction and risk stratification, where leveraging multi-modal data can provide a more comprehensive picture compared to relying solely on images.

\subsection{Foundational Models}

Recent advancements in foundational models for computer vision have produced exciting innovations. Vision transformers \cite{vit} and CNN-based baselines \cite{resnet} have continued to garner significant attention, and new baselines like \cite{swin,focalnet,vitdet} have established powerful baselines. Swin Transformer \cite{swin} challenges ViT's dominance by introducing hierarchical structures that improve efficiency for large image tasks. Focal Modulation Networks (FNet) \cite{focalnet} explore a different approach, utilizing a focus modulation mechanism to enhance the model's ability to attend to relevant features in complex scenes. Additionally, models like ViT-Det \cite{vitdet} demonstrate another way of adaptation of ViT for object detection tasks.

\section{Method}
\subsection{Revisiting Self-attention and Cross-attention}
Unlike traditional CNNs with limited local receptive fields, self-attention \cite{vit,vaswani2017attention} enables direct comparison of any two elements (regardless of their spatial distance) in the input, which is crucial for object detection by capturing relationships between distant regions:
\begin{align}
Attention(Q, K, V) = Softmax\left(\frac{QK^T}{\sqrt{d_k}}\right) V.
\end{align}
Here $Q, K, V$ are query, key, and value which is obtained from linear operation on features $X$, and $d$ is the dimension. Self-attention (when $Q$ and $V$ are generated from the same features) dynamically assigns weights to image regions based on task relevance, emphasizing important features like edges and suppressing background clutter, thereby enhancing object recognition. 
Similarly, cross-attention (when $Q$ and $V$ are generated from different features) extends model capabilities beyond the primary image, refining feature representations through interaction with learnable queries focused on specific aspects relevant to object detection, potentially yielding richer representations. 

\vita provides a specialised interaction block with the ViT Backbone, which consists of a convolution-based spatial feature pooling module, cross-attention-based feature injector and multi-scale extractor modules. 
%Unlike traditional CNNs, which are limited by local receptive fields, self-attention allows a model to directly compare any two elements in the input image, regardless of their distance. This is crucial for object detection, as understanding the relationships between different regions is essential for accurate object recognition. Additionally, self-attention \cite{vaswani2017attention} enables the model to dynamically assign weights to different image regions based on their relevance to the task. This allows focusing on important features like edges, shapes, and textures while suppressing background clutter. This focus improves the model's ability to identify and differentiate objects.
%On the other hand, cross-attention allows a model to incorporate information from external sources beyond the primary image. Cross-attention can be used to refine feature representations extracted by the backbone network. This can be achieved by interacting the features with learnable queries focused on specific aspects relevant for object detection, potentially leading to richer and more informative representations.

\subsection{Architecture of \myarch}
% As illustrated in the figure, our architecture can be viewed as combining three tiers: 1) A traditional pre-trained backbone; 2) The spatial prior injection and feature extraction modules based on ViT-Adapter\cite{}, 3) the learnable queries branch, which are initialised to zeros. The goal is to reduce the model's bias towards higher-level features and facilitate the flow of high-resolution features to the downstream tasks.
As illustrated in \cref{fig:arch}, our architecture consists of a ViT backbone, a spatial prior module, and an adapter branch containing a series of injectors, extractors, and learnable queries to help the model focus on regions of interest instead of spurious echogenic textures or noisy regions. Built on principles akin to ViT-Adapter, \myarch empowers ViT backbones (comprising Multiheaded Attention blocks) to excel in dense prediction tasks across datasets without necessitating architectural changes or re-training. The spatial prior module injects multi-scale features, enhancing localization-specific cues and refining features from pre-trained backbones. Injector and extractor blocks employ cross-attention to incorporate local spatial and image features from the backbone, supplementing missing information and reorganizing multi-scale features for dense predictions. However, due to the inherent simplicity of the spatial prior module, only using the CNN-based spatial priors were found insufficient for medical imaging tasks. Thus, we take inspiration from the object detection methods and integrate the learnable queries to learn richer object information. 
%\myarch works on the similar principles of ViT-Adapter and enables the ViT backbone to achieve competitive performance on dense prediction tasks across datasets without necessarily involving architectural changes or re-training. The spatial prior module injects multi-scale spatial features to provide additional vision-specific inductive biases and refine a pre-trained backbone. The injector and extractor blocks consist of cross-attention operations, which attend to the local spatial and image features from the backbone, which supplement missing local information and reorganize multi-scale features for dense prediction tasks. Since the backbone is frozen during adapter training, ViT-Adapter provides a flexible framework to adapt a backbone. 

\begin{figure*}[t]
    \centering
    \includegraphics[width=0.8\textwidth]{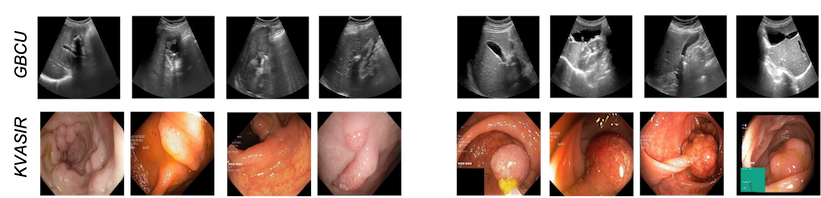}
    \caption{Sample images from GBCU \cite{basu2022surpassing}, and Kvasir-Seg \cite{kvasirseg} datasets. Malignant and benign samples from GBCU are on the left and right, respectively. Kvasir-Seg dataset \cite{kvasirseg} does not contain control images, so both sides showed images with polyp tissue}
    \label{fig:data_sample}
\end{figure*}

\subsection{Learnable Queries (LQ)}
To overcome the limitations of the spatial prior module, LQ-Adapter utilises learnable queries similar to those introduced in \cite{vqt,bao2021beit,Shehzadi2024SparseSS, Li2024SparseAT, Xiong2024MultimodalLQ, dab, dino,detrwithlcq}. The rationale is to couple information-rich ($F_{vit}^{i}$) features from the ViT backbone using attention. The learned queries reinforce the low-level features crucial to GBC detection, circumventing the limitations posed by the spatial prior module. The enhanced queries then engage in a cross-attention mechanism with the spatial prior module. This interaction refines the corresponding spatial features, enabling them to capture better the critical low-level information required for the next processing block.

DETR \cite{detr} introduced the use of learnable object queries (vector embeddings containing positional information) to help the decoder interact with the feature maps and help generate positional information for accurate localization of the objects in the image. %The flattened output of a conventional CNN was augmented with these queries before passing to the encoder block. % DAB DETR RELATED WORK SECOND PARA %
DAB-DETR \cite{dab} hypothesized that all information contained in queries are box coordinates and provided a framework to directly learn anchors as queries by extracting spatial features from a CNN backbone and feeding positional queries and decoder embeddings (acting as content queries) to their decoder. We employ a similar idea and introduce learnable queries, which are learned during the training process. %These allow the backbone to better attend to relevant localization information from the spatial embeddings and capture task-specific features effectively.
%
%Learnable content queries are implemented as tensors of similar size to the spatial priors, which are learned during the training process.
% Learnable content queries, similar in size to the spatial priors, are dynamically adjusted during training to focus the model's attention on informative features within the image.
Learnable queries are implemented as tensors matching the spatial priors' dimensionality, and are co-optimized via cross-attention with $F_{vit}^i$ \& $F_{sp}^{i+1}$ during training. These allow the model to better attend to relevant information from the spatial embeddings, allowing it to capture task-relevant features more effectively, without requiring a heavy spatial prior module, which would impair the lightweight nature of the architecture. 
%
% \todo{discuss what is learnable query? and the motivation to add it, add eqn of query if needed} - eq 3 and 4
%
%
% \subsection{ViT Backbone}:
% We use the traditional ViT Backbone, with L decoder layers, as the frozen backbone for our modelling purposes. Due to the frozen implementation, our trainable parameters drop significantly. The input for the ViT Backbone is from the spatial prior injection module of the second tier, which is then sent to the ViT Block for computing the corresponding ViT image features.

\subsection{Spatial Prior Injector}
Inspired from \vita, the primary function of the prior injector is to incorporate the priors into the backbone. To do so, the injector block consists of cross attention between input features from the backbone $F_{vit}^i$ and the spatial priors from the extractor block $F_{sp}^i$ to generate $\overline{F}_{vit}^i$. Here all $norm(.)$ are LayerNorm. 
\begin{align}
     \overline{F}_{vit}^{i} &= {F}_{vit}^{i} + \gamma^{i}*Attention(norm(F_{vit}^i), norm(F_{sp}^{i}))
\end{align}

\subsection{Extractor module with learnable queries}
Our extractor block consists of a cross attention between the output of the $i^{\text{th}}$ injector block $F_{sp}^i$ and the features from the backbone $F_{vit}^i \!\in\! {R}^{\frac{H\!\times\!W}{16^2}\!\times\!D}$, followed by a feed-forward network, to generate flattened multi-scale features $\overline{F}_{sp}^i \!\in\! R^{(\frac{HW}{8^2} + \frac{HW}{16^2} + \frac{HW}{32^2}) \times D}$ with scale 1/8, 1/16 and 1/32. Here, $H,W,\text{~and~}D$ are feature height, width, and depth, respectively. 
Additionally, cross attention is performed between zero-initialised learnable queries $LQ_i \in R^{\frac{HW}{16}\!\times\!D}$ and feature embeddings from our backbone module $F_{vit}^i$, which is further cross-attended with the multi-scale features to generate updated queries for the next block $LQ_{i+1}$. 
\begin{align}
    \overline{F}_{sp}^{i} &= Attention(norm(F_{vit}^i), norm(F_{sp}^{i})) \\
    F_{sp}^{i+1} &= \overline{F}_{sp}^{i} + FFN(norm(\overline{F}_{sp}^{i})) \\
    % LQ_{i} &= Attention(norm(\overline{F}_{vit}^{i}), norm(\overline{F}_{vit}^{i}))\\
    \overline{LQ}_{i} &= Attention(norm(LQ_{i}), norm(\overline{F}_{vit}^i))  \\
    LQ_{i+1} &= LQ_{i} + Attention(norm(F_{sp}^{i+1}),norm(\overline{LQ}_{i}))
\end{align}
Here $FFN$ and $norm$ refer to feed-forward network and normalization, respectively.
The introduction of learnable queries help the extractor block better attend to relevant information from the embeddings, allowing the architecture to capture task-relevant features more effectively, without using a heavy spatial prior module, which would impair the light-weight nature of the architecture. 
% \input{sections/table-detection}
% \subsection{Architectural Configuration}: 
% For training, we used a Uni-perceiver\cite{} backbone pre-trained on ImageNet-1k data. This backbone was frozen throughout the training process. We used AdamW optimizer with an initial learning rate of 6e-05 and a weight decay of 0.005. We also employ layer decay which decays the learning rate by a factor of 0.65 every 12 layers. All injector and extractor blocks were trained for 60 epochs with a batch size of 2. We also used center cropping and normalization before training to avoid overfitting. 

\section{Datasets}
\begin{table*}[t]
	\centering
	\setlength{\tabcolsep}{8pt}
	\caption{The detection/ localization performance comparison of our method and SOTA baselines for the GBCU dataset. FocalNet-DINO We report mIoU, precision, and recall. }
    %\vspace{0.5em}
   % \resizebox{\textwidth}{!}{%
	\begin{tabular}{lccc}
		\toprule
		\textbf{Method}	& \textbf{mIoU} & \textbf{Precision} & \textbf{Recall} \\
		\midrule
		%FasterRCNN \cite{fasterrcnn} & 0.711 $\pm$ 0.027 & 0.960 $\pm$ 0.026 & 0.992 $\pm$ 0.007 \\
        %    YOLOv4 \cite{bochkovskiy2020yolov4} &  0.707 $\pm$ 0.029 & 0.981 $\pm$ 0.023 & 0.979 $\pm$ 0.015 \\
        %    CentripetalNet \cite{dong2020centripetalnet} & 0.604 $\pm$ 0.047 & 0.951 $\pm$ 0.038 & 0.896 $\pm$ 0.073 \\
		%\midrule
		% \multirow{4}{*}{Foundational Models} 
            DINO \cite{dino}  & 0.661 $\pm$ 0.032& 0.991 $\pm$ 0.010 & 1.000 $\pm$ 0.000 \\
		FocalNet-DINO \cite{focalnet}  & 0.692 $\pm$ 0.014 & 0.994 $\pm$ 0.005 & 0.998 $\pm$ 0.002\\
            ViT-Adapter \cite{vita}  & 0.665 $\pm$ 0.020 & 0.972 $\pm$ 0.016 & 0.999 $\pm$ 0.002 \\
		% \cmidrule{2-5}
            \midrule
		  \myarch & 0.719 $\pm$ 0.021 & 0.981 $\pm$ 0.008 & 0.999 $\pm$ 0.004 \\
		\bottomrule
	\end{tabular}
	% }
	\label{tab:detection}
\end{table*}
\begin{table*}[t]
	\centering
	\setlength{\tabcolsep}{8pt}
	\caption{The performance comparison of classifying malignant vs. non-malignant GBs from US images. We report the accuracy, specificity, and sensitivity. GBCNet and Radformer were the previous SOTA for the task. Augmenting the GBCNet architecture with using \myarch as the ROI generator improves the GBC classification accuracy notably over the previous SOTA.}
    %\vspace{0.5em}
    %\resizebox{ 0.8\textwidth}{!}{%
	\begin{tabular}{lccc}
		\toprule
		\textbf{Method}	& \textbf{Acc.} & \textbf{Spec.} & \textbf{Sens.} \\
		\midrule
		% ViT \cite{vit} &  0.803 $\pm$ 0.078 & 0.901 $\pm$ 0.050 & 0.860 $\pm$ 0.068\\
		%
		% DEIT\cite{deit} & 0.829 $\pm$ 0.030 & 0.900 $\pm$ 0.040 & 0.875 $\pm$ 0.063 \\

            % PVTv2\cite{pvt} & 0.824 $\pm$ 0.033 & 0.887 $\pm$ 0.057 & 0.894 $\pm$ 0.076 \\
            
            % LoRA-ViT \cite{loravit} & 0.75 & 0.61 & 0.825 \\
            
            % EDEN \cite{eden} & 0.80 & 0.925 & 0.574 \\
		%
		% EFFT \cite{efft} & 0.86 & 0.975 & 0.64 \\
		%
            ViT \cite{vit} & 0.803$\pm$0.078 & 0.901$\pm$0.050 &  0.860$\pm$0.068 \\
		Radformer \cite{basu2023radformer} (Prev. SOTA) & 0.921 $\pm$ 0.062 & 0.961 $\pm$ 0.049 & 0.923 $\pm$ 0.062 \\
            GBCNet (w/ Faster-RCNN ROI) \cite{basu2022surpassing} & 0.861 $\pm$ 0.087 & 0.867 $\pm$ 0.098 & 0.844 $\pm$ 0.097 \\
            \midrule
            GBCNet (w/ DINO ROI) & 0.886 $\pm$ 0.020 & 0.889 $\pm$ 0.020 & 0.853 $\pm$ 0.040 \\
            GBCNet (w/ FocalNet-DINO ROI) & 0.882 $\pm$ 0.020 & 0.889 $\pm$ 0.020 & 0.853 $\pm$ 0.040 \\
		\midrule%[1.5pt]
	   GBCNet (w/ \myarch ROI) & \textbf{0.934 $\pm$ 0.022} & 0.938 $\pm$ 0.026 & \textbf{0.923$\pm$ 0.028}\\
		\bottomrule
	\end{tabular}
	%}
	\label{tab:classification}
\end{table*}
\subsection{Gallbladder Cancer Ultrasound Data}
We use the publicly available GBCU dataset \cite{basu2022surpassing}, which is suitable for both classification and detection, to assess the GBC detection performance of the proposed \myarch. GBCU comprises 1255 B-mode Ultrasound (US) images of the Gallbladder (GB) from transabdominal scans, including 265 malignant and 990 non-malignant images collected from 171 non-malignant and 47 malignant patients. Each anonymized image ranges in width from 801 to 1556 pixels and in height from 564 to 947 pixels. Additionally, each image includes a region-of-interest (ROI) delineating the GB and pathology, marked with an axis-aligned bounding box. The dataset provides a default train and validation split, with 1133 and 122 images, respectively, while we report 5-fold cross-validation results to address biases in the small dataset. We use patient-level cross-validation splits to ensure data integrity, with all images of any patient appearing in either the train or validation split.

\subsection{Kvasir-Seg Polyp Detection Data}
Additionally, we use Kvasir-Seg \cite{kvasirseg}, which is a publicly available dataset designed to address the challenge of limited annotated data in gastrointestinal polyp segmentation, to show the generality of our method. %The data  a gastroenterologist, ensuring pixel-wise accuracy. 
%The data contains bounding boxes for the polyp regions. %were generated based on the segmentation masks. 
The Kvasir dataset provides 1000 annotated colonoscopy images containing polyps, and the corresponding bounding boxes and segmentation masks for the polyp region. Image dimensions vary between 352 to 1072 pixels in height and 332 to 1920 pixels in width. 

%DDSM \cite{ddsm} is a publicly available dataset containing mammograms obtained from different views of each breast, along with ground truth labels wherein each mammogram is classified as either normal or abnormal. In case of abnormalities, the type and location of the abnormality is also present in the dataset. DDSM contains 3042 samples out of which 1604 are normal and 1438 samples contain abnormalities. The height of images range between 3256 to 6962 pixels and the width ranges from 1510 to 5394 pixels. For our experiments, we split the dataset into train, test and val sets with 2166, 639 and 237 image samples, respectively. The samples contain bounding box annotations of 1438 abnormalities. The height of annotations range between 35 and 3779 pixels, while the width ranges from 42 to 2876 pixels. %\todo{add details of obj det labels} - done
%
% \todo{fill ddsm details}

%-------------------------------------------------------------------------

\begin{table*}[t]
	\centering
	\setlength{\tabcolsep}{10pt}
	\caption{The detection/ localization performance comparison of our method and SOTA baselines for the Kvasir dataset \cite{kvasirseg} for polyp detection in Colonoscopy. We report mIoU, precision, and recall. The consistent performance improvement of \myarch across both GBC and polyp detection indicates generality of our method.}
    %\vspace{0.5em}
    %\resizebox{ \linewidth}{!}{%
	\begin{tabular}{lcccc}
		\toprule
		\textbf{Method}	& \textbf{mIoU} & \textbf{Precision} & \textbf{Recall} \\
		\midrule
        DINO \cite{dino}  & 0.848 & 0.966 & 1.0 \\
        FocalNet - DINO\cite{focalnet}  & 0.85 & 0.984 & 1.0 \\
        ViT-Adapter \cite{vita} & 0.812 & 0.975 & 1.0  \\
        \midrule
    	Ours (\myarch) & \textbf{0.85} & 0.966 & 1.0 \\   
    	\bottomrule
	\end{tabular}
	%}
	\label{tab:ddsm}
\end{table*}

\section {Experiments and Results}
\subsection{Experimental Setup} 
We use a machine with Intel Xeon Gold 5218@2.30GHz processor and 4 Nvidia Tesla V100 GPUs for our experiments. We used a Uni-perceiver \cite{uniperceiver} backbone pre-trained on ImageNet-1k data \cite{imagenet}. Note that the backbone was frozen throughout the training process, and only the adapter block was trained. We used AdamW optimizer with an initial learning rate of 6e-05 and a weight decay of 0.005. We also employ layer decay which decays the learning rate by a factor of 0.65 every 12 layers. All injector and extractor blocks were trained for 60 epochs with a batch size of 2. We also use data augmentations such as center cropping and normalization. 

\subsection{Evaluation Metrics}
%updatee point to gbc 
Building upon prior work in Gall Bladder Cancer (GBC) detection \cite{basu2022surpassing, basu2023radformer, gbc-xgc}, we employ a comprehensive suite of metrics to evaluate our network's performance in object detection.  Mean Intersection-over-Union (mIoU) serves as the primary metric for object detection. It measures the average overlap between predicted bounding boxes and ground truth annotations. Additionally, we assess localization performance using precision and recall. Localization precision and recall are calculated following \cite{ribli2018detecting}, where a region prediction is considered true positive if its centre lies within the ground truth bounding box; otherwise, it's considered false positive due to localization error, whereas no predictions are marked as false negatives. 

To evaluate the model's ability to correctly classify the presence or absence of GBC, we utilize accuracy, specificity, and sensitivity. Accuracy reflects the overall rate of correctly classified images. Specificity measures the class-wise accuracy of benign samples (True Negative Rate), while sensitivity measures the class-wise accuracy of detecting malignancy (True Positive Rate/ Recall).
% Classification performance is assessed using the accuracy, specificity, and sensitivity of class label predictions.

%We use the mean intersection-over-union (mIoU), localization precision and recall for assessing the detection performances of the networks. For calculating the localization precision and recall we follow the method suggested by \cite{ribli2018detecting}. If the center of the predicted region lies within the bounding box of the ground truth region, then we consider a region prediction to be a true positive; otherwise, we consider the region prediction to be a false positive due to localization error. Further, we consider the zero/ no region prediction as a false negative. For assessing the classification performance, we use the accuracy, specificity, and sensitivity of the class label predictions.

% \input{sections/table-detection}
% \input{sections/table-classification}
% \input{sections/table-ddsm}

%\input{table-detection}
\subsection{Comparison with SOTA Detectors}
\begin{figure*}[t]
    \centering
    \includegraphics[width=0.9\textwidth]{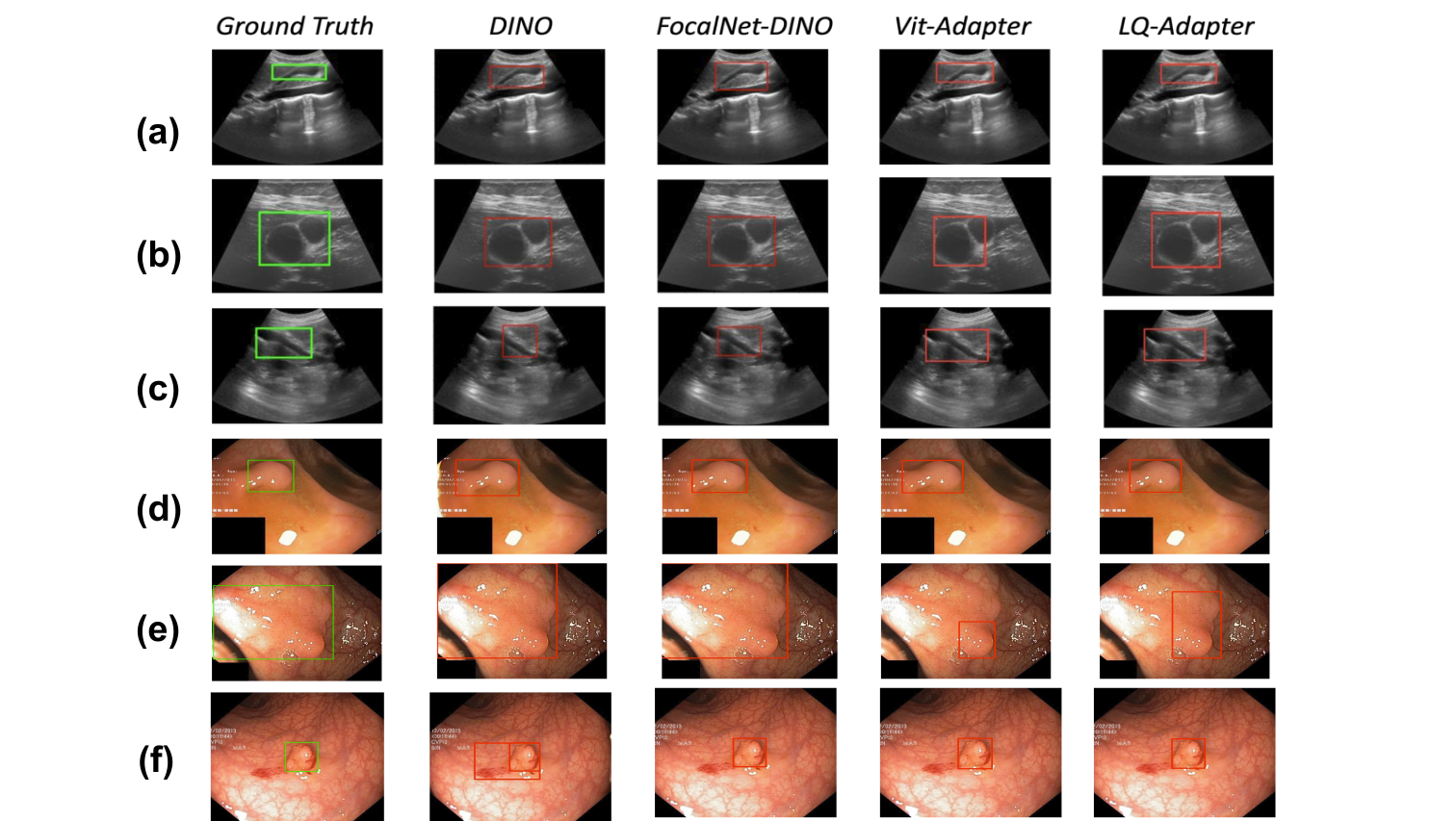}
    \caption{We motivate the use of learnable content queries in the adapter design. We show sample localizations by \vita \cite{vita}, DINO \cite{dino}, FocalNet-DINO \cite{focalnet, focalstabledino}, and LQ-Adapter (ours). Primitive spatial prior modules in \vita do not capture the salient region well, reducing detection quality. \myarch, on the other hand, can learn the region information well via the learnable queries and thus demonstrate superior localization performance. Rows (a)-(c) show selected samples from the GBCU dataset \cite{basu2022surpassing} and rows (d)-(f) are samples from the Kvasir-Seg dataset \cite{kvasirseg}. (green bounding boxes: ground truth, red bounding boxes: prediction).}
    \label{fig:qualitative}
\end{figure*}

%
%  expand
In \cref{tab:detection}, we compare the object detection performance of our proposed \myarch on GBCU dataset with the SOTA object detectors. Please note that Focal Modulation Backbone-based DETR \cite{focalnet, focalstabledino} currently beats all the existing object detection baselines, eliminating the need for comparison against older baselines. For GBCU, the mIoU of our model outperforms \vita, DETR-based SOTA (namely DINO and FocalNet-DINO) detectors. The proposed \myarch surpasses all SOTA detection methods, thus establishing a new SOTA. \cref{fig:qualitative} shows predicted bounding box visuals for baselines and \myarch. 

Another clear advantage our model offers is in terms of the number of trainable parameters. \cref{fig:teaser} Specialised detectors like DINO (Swin variant) and FocalNet-DINO (FocalNet-Large variant) have trainable parameters in the order of  228,053,892 (230 Million+). This calls forth two disadvantages; firstly, for small medical datasets, these models often tend to overfit, leading to worse performance on unseen data. Secondly, fine-tuning these models requires heavy computing resources, making re-training the models with new data cumbersome. \myarch and \vita have 56\% of the trainable parameters (about 135,530,325(130 Million)) and still produce comparable or better mean-intersection-over-union for the GBCU dataset. This observation continues to hold true over the Kvasir-Seg \cite{kvasirseg} dataset, with \myarch showing a comparable performance over the FocalNet-DINO DETR and DINO-DETR \cite{dinodetr}.    %We also show in supplementary \cref{fig:qualitative} additional visualizations of the detection results with \myarch. 
Compared to the SOTA DETR variants, or the \vita, we achieve superior localization.

\subsection{Comparison for GBC Classification}
Additionally, for the classification of malignant and non-malignant GBs, existing literature suggests that using focused regions helps mitigate the effect of noise and artefacts in US images and helps improve GBC classification performance \cite{basu2022surpassing}. Previously, GBCNet \cite{basu2022surpassing} used Faster-RCNN-based candidate region generation, followed by a specialized classifier head called Multi-Scale Second-Order Pooling (MS-SoP). Our work builds upon this concept by introducing \myarch as a novel candidate region generation method.  We integrated \myarch into the GBCNet architecture, replacing the Faster R-CNN component.  This substitution resulted in a significant improvement of 2.3\% in classification accuracy, as shown in \cref{tab:classification}. 

Furthermore, GBCNet with \myarch also outperformed the current state-of-the-art (SOTA) method, RadFormer \cite{basu2023radformer}, by 1.3\%. These impressive results highlight the effectiveness of \myarch in pinpointing the relevant regions within the ultrasound image, ultimately leading to more accurate GBC classification. We show these results in \cref{tab:classification}. The superior performance with using \myarch as the region-of-interest (ROI) generator reinstates its efficacy as a GB malignant region localizer. 

\begin{figure}[t]
    \centering
    \begin{subfigure}[b]{0.5\linewidth}
        \centering
        \includegraphics[width=\linewidth]{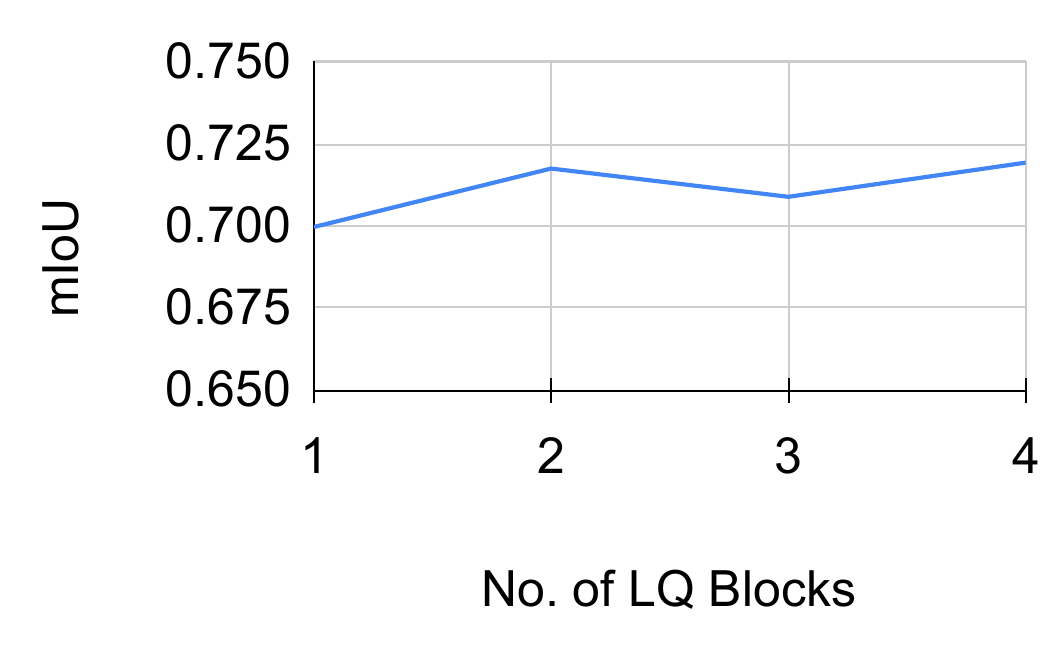}
        \caption{}
        \label{fig:ablation1}
    \end{subfigure}
    %
    %\qquad
    \begin{subfigure}[b]{0.4\linewidth}
        \centering
        \includegraphics[width=\linewidth]{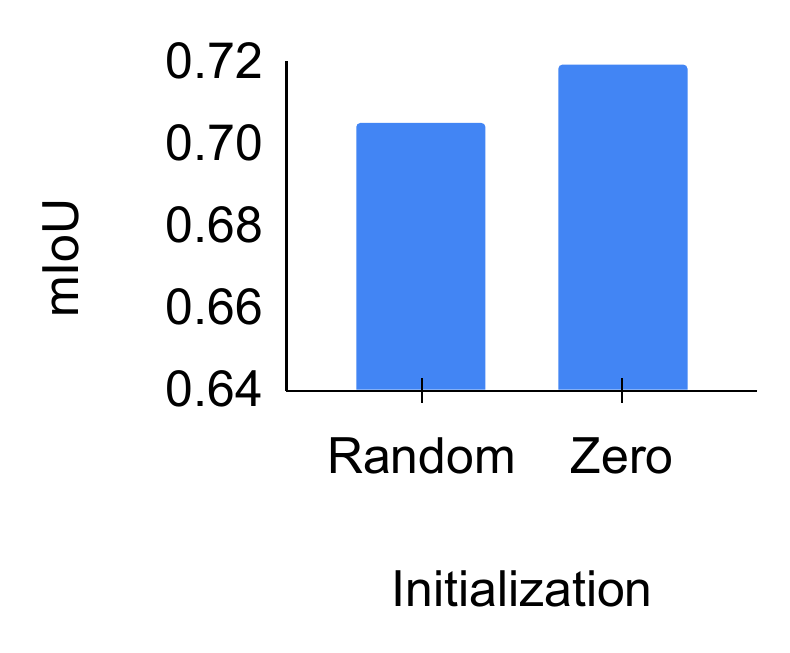}
        \caption{}
        \label{fig:ablation2}
    \end{subfigure}
    \caption{Ablation Study. (a) Shows the effect of the number of learnable query blocks on performance. We observe that augmenting all layers with learnable queries results in the highest performance gain. (b) The effect of initializing the queries with zero values and random values.}
\end{figure}

\subsection{Evaluating Generalizability}
We assess the generality and applicability of the proposed \myarch on the task of polyp detection from colonoscopy images. We use the Kvasir-Seg dataset \cite{kvasirseg} for polyp detection. We report the results in \cref{tab:ddsm}. Our proposed model outperforms \vita for polyp detection by 4.1\% in terms of mIoU. Further analysis shows that the results hold comparable performance against FocalNet-DINO and DINO DETR\cite{dinodetr, focalnet, focalstabledino} while maintaining a significantly lower parameter count with approximately 56.5\% fewer parameters. 
%    
%     \item \textbf{DDSM dataset} \cite{ddsm} for breast cancer detection from mammograms: As shown in \cref{tab:ddsm}, our proposed model outperforms \vita for breast lesion detection by 14.3\% in terms of mIoU.
% \end{itemize}
%
\par These results on two distinct tasks -- (1) GBC detection in ultrasound images and (2) Gastrointestinal polyp detection from colonoscopy images -- provide compelling preliminary evidence that \myarch generalizes well across different medical image modalities and disease types. This broad applicability suggests the potential of \myarch as a versatile tool for various medical image analysis applications.

\subsection{Ablation Study}
\mypara{Choice of the block for LQ} 
We compared the performance of detection with LQ introduced at different blocks on the GBCU dataset in \cref{fig:ablation1}. Increasing the number of LQ blocks positively impacts the performance. LQ at all the blocks shows the best performance.

\mypara{Choice of initialization for LQ}
The detection performance was affected by the choice of initialization of the queries. We experimented with random initialization and initialization from zero. As seen in \cref{fig:ablation2}, using zero initialization results in a better performance gain.

\subsection{Qualitative Results and Discussion}

We have showcased a qualitative comparison between the existing baselines and \myarch in \cref{fig:qualitative}. %Regarding ultrasound image analysis, the \myarch stands out for its exceptional qualitative strengths. Let's delve into the key advantages it offers:

%\mypara{Holistic Feature Capture}
%
\myarch excels at capturing comprehensive features within ultrasound images and localizing the entire gallbladder and its surrounding area, including the pathology. 
%By capturing the whole picture, \myarch provides a complete understanding of the anatomy.

\mypara{Discerning Subtle Details} A common observation with \myarch is its ability to catch clinically crucial markers in the case of GBC. As noticeable in \cref{fig:qualitative}(a,b,c), \myarch meticulously identifies the exact boundaries of the gallbladder and neighbouring hypoechoic regions. LQ-Adapter's approach often captures slightly larger areas around the target object than the bounding boxes captured by other methods. This is advantageous for medical image analysis as it encompasses relevant information about the task. For instance, in evaluating GBC, involving dense tissue around the edges might provide clues about potential abnormalities that would be easily missed with a tighter bounding box.

\mypara{Additional Analysis} Qualitative analysis on the Kvasir-Seg dataset \cite{kvasirseg} further strengthens the case for \myarch's generalizability. In a head-to-head comparison with \vita, \myarch consistently aligns better with radiologist annotations. \vita produces tighter bounding boxes, which can miss important clinical markers like the surrounding polyp tissue. Additionally, while the high intra-class variability (extent of polyp spread in the Kvasir-Seg dataset) poses a challenge for \vita, \myarch is able to adapt better to these situations as seen in \cref{fig:qualitative}(d,e). Finally, DETR variants identifying noisy regions in (e), and multiple polyps in (f) raises concerns for false positive detections, which can undermine diagnostic trust. While our model performance is similar to that of FocalNet DINO in the case of this dataset, we would like to re-iterate attention to the significant advantage \myarch offers in terms of trainable parameters (half as compared to FocalNet DINO).

\mypara{Downstream Applications} Lastly, the higher accuracy reflected in agreeing with radiologist annotations makes it ideal for generating spatial priors. Spatial priors act as knowledge maps for advanced models like those described in \cite{basu2023radformer} and \cite{focusmae}. By providing these advanced models with precise spatial information about the expected locations and relationships between anatomical structures, \myarch lays the groundwork for more accurate diagnoses \& predictions.

\section{Conclusion}
This paper introduces \myarch, a novel adapter with learnable queries, enabling ViT for object detection/localization in medical imaging tasks, particularly Gallbladder Cancer (GBC) detection from US images. Achieving SOTA performance without task-specific architectures, \myarch also extends applicability to breast lesion detection. By showcasing the effectiveness of foundational models, we aim to spark interest in leveraging similar approaches for diverse medical imaging tasks.
%In this paper, we have developed \myarch, a novel adapter with learnable queries, for enabling the ViT for object detection/ localization on medical imaging tasks. We open the research avenues for effectively leveraging foundational architectures for GBC detection from US images. Our proposed \myarch achieves SOTA performance for GBC detection, bypassing the need for developing specific architectures previously designed towards the task. Our proposed model is applicable on breast leasion detection as well. We hope that the model would generate interest in the community in the direction of effectively designing foundational architectures for medical imaging tasks.

%%%%%%%%% REFERENCES
{\small
\bibliographystyle{ieee_fullname}
\bibliography{main}
}

\end{document}